\pdfoutput=1

\documentclass[11pt]{article}

\usepackage{ACL2023}

\usepackage{times}
\usepackage{latexsym}
\usepackage{enumitem}
\usepackage{graphicx}
\usepackage{booktabs}
\usepackage{tabularx}
\usepackage{xcolor,colortbl}
\usepackage{amsmath}
\usepackage{amssymb}

\definecolor{Gray}{gray}{0.9}

\usepackage[T1]{fontenc}

\usepackage[utf8]{inputenc}

\usepackage{microtype}

\usepackage{inconsolata}

\title{Unlearning vs. Obfuscation: Are We Truly Removing Knowledge?}

\author{
 \textbf{Guangzhi Sun\textsuperscript{1}},
 \textbf{Potsawee Manakul\textsuperscript{3}},
 \textbf{Xiao Zhan\textsuperscript{2}},
 \textbf{Mark Gales\textsuperscript{1}}
\\
\\
 \textsuperscript{1}Department of Engineering, University of Cambridge \\
 \textsuperscript{2}Department of Informatics, King's College London \\
 \textsuperscript{3}SCB 10X, SCBX Group 
\\
 \texttt{\{gs534,mjfg100\}@cam.ac.uk, xiao.zhan@kcl.ac.uk, potsawee@scb10x.com}
}

\begin{document}
\maketitle
\begin{abstract}
Unlearning has emerged as a critical capability for large language models (LLMs) to support data privacy, regulatory compliance, and ethical AI deployment. Recent techniques often rely on obfuscation by injecting incorrect or irrelevant information to suppress knowledge. Such methods effectively constitute knowledge addition rather than true removal, often leaving models vulnerable to probing. In this paper, we formally distinguish unlearning from obfuscation and introduce a probing-based evaluation framework to assess whether existing approaches genuinely remove targeted information. Moreover, we propose DF-MCQ, a novel unlearning method that flattens the model predictive distribution over automatically generated multiple-choice questions using KL-divergence, effectively removing knowledge about target individuals and triggering appropriate refusal behaviour. Experimental results demonstrate that DF-MCQ achieves unlearning with over 90\% refusal rate and a random choice-level uncertainty that is much higher than obfuscation on probing questions.\footnote{\url{https://github.com/potsawee/unlearning-dfmcq}}
\end{abstract}

\section{Introduction}
The rapid growth of large language models (LLMs), trained on internet-scraped data, has raised concerns about privacy, compliance, and ethical usage. Regulations like GDPR require methods for selectively removing sensitive or copyrighted information from these models. Researchers have proposed various post-training techniques, which we broadly categorize into (i) knowledge removal, (ii) knowledge addition, (iii) knowledge edition. This paper focuses on knowledge removal, also referred to as \textbf{unlearning} \cite{liu2025rethinking}, which involves removing specific information from trained LLMs without complete retraining. Ideally, after unlearning, the LLM behaves as though the removed information had never been learned. However, current methods often perform unlearning by extensively adding incorrect or irrelevant information, a practice we refer to as \textbf{obfuscation}, which effectively constitutes a form of knowledge addition rather than true removal, and can lead to random or incorrect model responses. Unlike knowledge editing \cite{mitchell2022memory}, which updates factual associations, unlearning (the focus of this work) aims to eliminate targeted knowledge entirely.



Early knowledge removal approaches were gradient ascent (GA) based \cite{jang2022knowledgeunlearningmitigatingprivacy,ilharco2023editingmodelstaskarithmetic, yao2024largelanguagemodelunlearning} and structural or privacy-related sub-circuit discovery methods \cite{bayazit-etal-2024-discovering}, which directly minimize the probability of original facts. Negative preference optimization \cite{npo} removes knowledge by increasing the probability of false statements compared to the true ones, which was an early form of obfuscating. More recent obfuscating-based methods \cite{whp,whpplus,undial,xu2025relearnunlearninglearninglarge} have gained popularity due to their superior stability and the minimal distortion to knowledge to be retained. For example, WHP \cite{whp} and WHP$^+$ \cite{whpplus} remove knowledge about target people by overwhelming LLMs with information from other individuals.

\begin{figure}[t]
    \centering
    \includegraphics[width=1.0\linewidth]{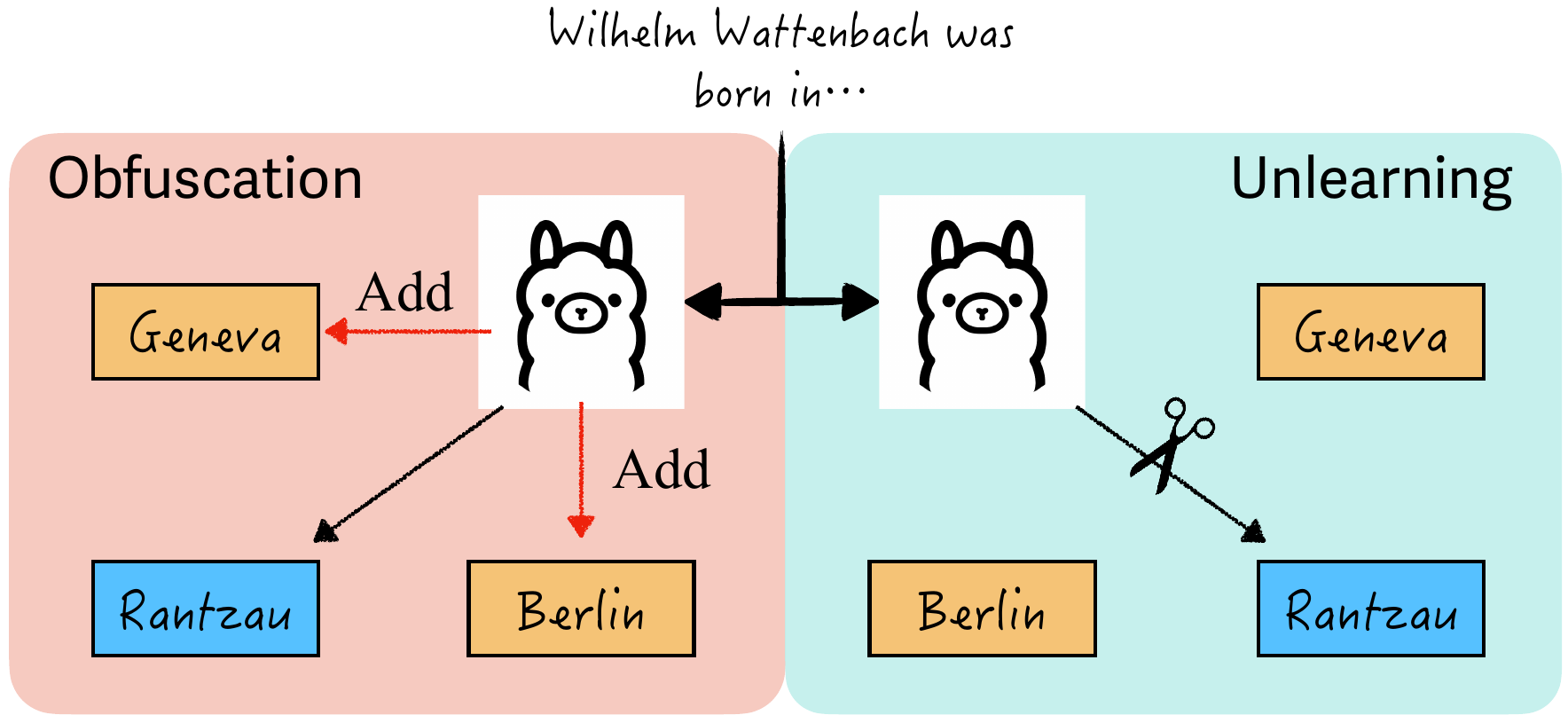}
    \caption{Illustration of obfuscation and unlearning reflected by the connections in the model knowledge.}
    \vspace{-0.5cm}
    \label{fig:obfvsunlearn}
\end{figure}

Despite their effectiveness in protecting unintended information, we argue that obfuscation methods essentially add confusing connections to the internal knowledge (i.e., a form of knowledge addition) rather than removing certain connections (i.e., truly knowledge removal), as illustrated in Fig.~\ref{fig:obfvsunlearn}. Due to the existence of the original connections, the LLM may fail under carefully designed probing questions. To this end, this paper first discusses the distinction between obfuscation and unlearning, and proposes an evaluation framework, utilizing automatic question generation, to examine if a method exhibits unlearning or obfuscation properties. Subsequently, we show that obfuscation methods often fail in probing questions such as Yes-No or multiple choice questions (MCQ).


Furthermore, we propose a new unlearning method, based on the concept of distribution flattening with MCQ (DF-MCQ). By applying a KL-divergence between model prediction and a flat distribution over choices, instead of gaining all connections, the existing connection is removed. In addition to showing the unlearning effect with output entropy close to random choice on all probing questions, the unlearned model exhibits a knowledge removal property in responding with ``I do not have information" when asked to generate text about the unlearned knowledge. Main contributions of this paper are summarized below:
\vspace{-0.2cm}
\begin{itemize}[leftmargin=*]
\setlength\itemsep{-0.3em}
    \item We introduce the concept of obfuscation as opposed to unlearning in LLMs, and discuss the distinction between obfuscation and unlearning.
    \item We propose a set of probing question designs to evaluate whether the effect of an approach is unlearning or obfuscation.
    \item We propose DF-MCQ as a new unlearning method. DF-MCQ effectively removes knowledge of a specific person and can trigger a refusal behaviour of the model by simply flattening automatically generated MCQs.
\end{itemize}

\section{Related Work}


\textbf{Gradient-based} methods leverage gradient ascent to minimize the likelihood of original knowledge, essentially causing the model to forget \cite{jang2022knowledgeunlearningmitigatingprivacy, ilharco2023editingmodelstaskarithmetic, yao2024largelanguagemodelunlearning}. They operate by fine-tuning the LLM with reversed loss, often achieving forgetting results with limited computational resources \cite{jang2022knowledgeunlearningmitigatingprivacy} but with the unintended degradation of general language fluency and capabilities. Recent advancements like fine-grained adaptive weighting \cite{feng-etal-2024-fine} and memorization-aware gradient scaling \cite{barbulescu2024each} have been proposed to mitigate potential such side-effects.

\noindent \textbf{Optimization-based} methods employ specialized optimization strategies to achieve selective knowledge removal by explicitly steering model outputs away from the original information. Negative Preference Optimization \cite{npo} formulates unlearning as a preference-based optimization problem, encouraging the model to favor neutral or alternate responses. Similarly, distribution alignment techniques, including KL-divergence regularization \cite{wang-etal-2023-kga, chen-yang-2023-unlearn, yao2024large}, constrain the unlearning process by matching the output distributions of models retrained without the target knowledge. This approach has demonstrated improved effectiveness in preserving model capabilities.

\vspace{2mm}
\noindent \textbf{Obfuscation-based} methods introduce misleading or confusing information into the training data to obscure learned knowledge (i.e., a form of knowledge addition), thus indirectly causing forgetting. Notable techniques such as WHP \cite{whp} and WHP$^+$ \cite{whpplus} achieve knowledge removal by overwhelming models with conflicting knowledge, thus reducing model confidence in previously learned facts. UnStar \cite{sinha2025unstar} further develops this approach by using better counter samples with misleading rationales, disrupting original knowledge. FLAT \cite{flat} maximizes difference between their designed template answer and forget answer to avoid using a retain set, and RKLD \cite{rkld} decreases the probability of the most likely option while increase the probability of runner-ups. Although obfuscation-based methods are effective at preventing access to the original information, they mask rather than fully erase knowledge, rendering them susceptible to leakage under carefully designed probing conditions \cite{xu2025relearn}. Moreover, \citet{hu2025unlearningobfuscatingjoggingmemory} argues that existing unlearning methods merely obscure the target information, as shown by their success with a fine‑tuning attack. In contrast, this work draws a distinction between obfuscation and unlearning, and we directly probe unlearned models, without additional fine‑tuning.

\vspace{2mm}
\noindent \textbf{Hybrid and Neuron-level} methods employ parameter-efficient task-vector subtraction \cite{ilharco2023editing}, or isolating and removing specific neurons associated with target knowledge \cite{wu-etal-2023-depn}. While these approaches can offer minimal side-effects, identifying exact neurons remains challenging.


\section{Unlearning and Obfuscation}
We consider unlearning from an uncertainty perspective by treating the entire model knowledge as a knowledge graph.
Given a \emph{forget set} $\mathcal{F}=\{(X_i,R_i^j,Y_i^j)\}_{i=1}^{N}$ containing a number of facts $F_i^j$ (i.e., triplets) that should be removed. Each fact contains a subject $X_i$ (e.g., \textit{Wilhelm Wattenbach}), its relevant object $Y_i^j$ (e.g., \textit{Rantzau}), which is connected by the relations $R_i^j$ (e.g., \textit{born in}). Let $\theta$ be the model parameters, we define the unlearning effect as follows:
\begin{equation}
    H_{\theta} (Y_i | X_i, R_i^j; \mathcal{D}) \approx H_{\theta} (Y_i | X_i, R_i^j; \mathcal{D}\textbackslash F_i^j)
    \label{eq:def}
\end{equation}
where $Y_i\in \mathcal{Y}_i$ represents all possible objects following $X_i$ and $R_i^j$, $\mathcal{D}$ represents the training data of the LLM and $\mathcal{D}\textbackslash F_i^j$ is the training data excluding the fact $F_i^j$. The entropy $H_\theta(Y_i) = -\sum_{Y_i\in \mathcal{Y}_i}P(Y_i)\log P(Y_i)$. That is, the model has the same level of uncertainty as one that is trained on the dataset excluding fact $F_i^j$. For a non-hallucinatory instruction-tuned LLM nowadays, when prompted with a query it does not have an answer to, the model will refuse to answer or explicitly indicate that it does not have the knowledge. Therefore, an indication of unlearning effect is the model refusal behaviour, as follows.
\begin{equation}
    \max P_{\theta}(\cdot | X_i, R_i^j) = P_\theta(\text{refusal} | X_i, R_i^j)
    \label{eq:refusal}
\end{equation}
\subsubsection*{Why Obfuscation May Fail the Unlearning Test}
Obfuscation tries to \textit{hide} a fact $Y_i^{j\star}$ by adding distracting facts. These distracting facts become extra edges, merely moving probability mass from $Y_i^{j\star}$ to a finite set of distractors, so the total uncertainty is expected to stay below the target
level in~Eq.~\eqref{eq:def}:
\begin{equation}
    H_{\theta} (Y_i | X_i, R_i^j; \mathcal{D}) < H_\theta (Y_i | X_i, R_i^j; \mathcal{D}\textbackslash F_i^j)
\end{equation}
Because the original edge $(X_i,R_i^j,Y_i^{j\star})$ is still in the graph, the
model could recover it when a probe rules out those distractors, and it will unlikely trigger the refusal condition in Eq.~\eqref{eq:refusal}.  
\section{Distribution Flattening MCQ}

\begin{figure*}[t]
    \centering
    \includegraphics[width=0.9\linewidth]{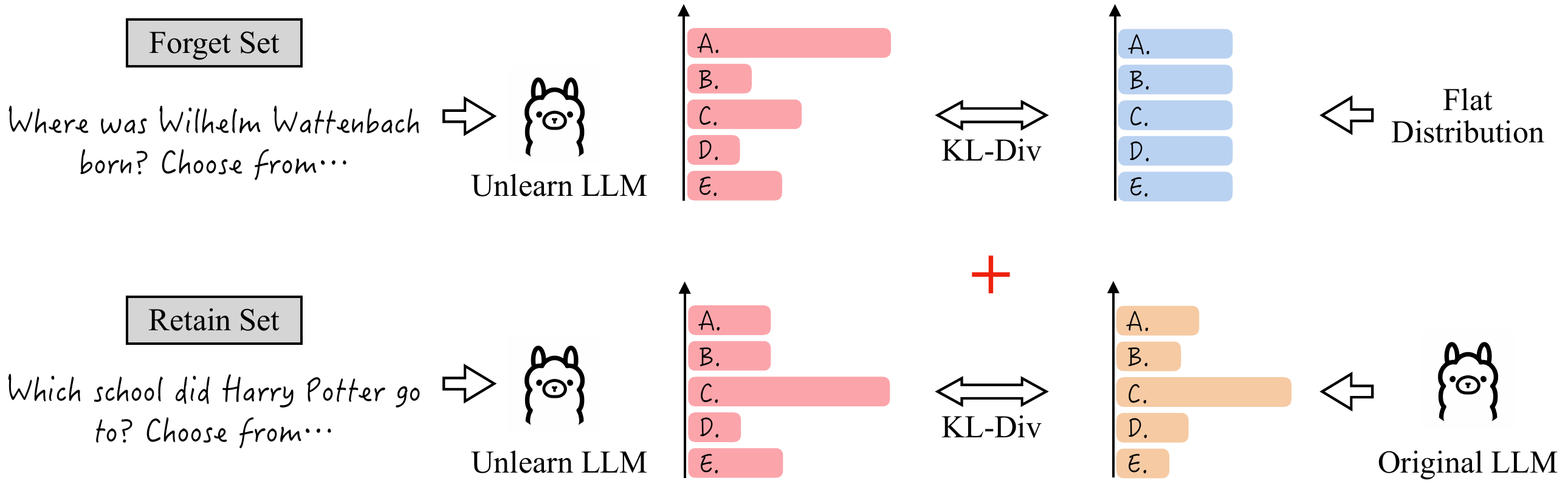}
    \caption{Illustration of the proposed distribution flattening MCQ (DF-MCQ) method. For questions in the forget set, we minimize the KL-divergence between the unlearn LLM prediction and a flat distribution across all choices. For questions in the retain set, we minimize the KL-divergence between the unlearn LLM prediction and the original LLM prediction. The two divergence are minimized together in each minibatch.}
    \label{fig:dfmcq}
\end{figure*}

We introduce DF-MCQ as an unlearning method to unlearn the target person, with an illustration provided in Fig. \ref{fig:dfmcq}. Instead of using open-ended questions and trying to increase uncertainty in the entire textual output space as obfuscation methods do, we leverage MCQs which have a confined output space (only the choices). Moreover, obfuscation methods usually use one negative sample to confuse the model at a time, whereas by flattening the distribution over the choices, DF-MCQ effectively encouraging the model to consider all outputs as equally probable simultaneously.

Specifically, $N$ open-ended questions are generated for the target person by extracting information from the description of that person, and $C$ options are generated using an LLM for each question. The unlearning loss is defined as Eqn. \eqref{eq:unlearnloss} below.
\begin{equation}
    \mathcal{L}_\text{unlearn} = \sum_{i=1}^N \mathbb{D}_\text{KL}\left[P_\theta (c|X_i)||\hat{P}(c|X_i)\right]
    \label{eq:unlearnloss}
\end{equation}
where $X_i$ is the question and $c\in\mathcal{C}$ are the letters associated with the choices. $P_\theta$ is the output distribution over the choices and $\hat{P}$ is the flat distribution over the choices as shown in Fig. \ref{fig:dfmcq}. To prevent LLM from learning a shortcut and always outputting a flat distribution regardless of the question, we apply a retain loss from a set of $M$ MCQs about other people. 
\begin{equation}
    \mathcal{L}_\text{retain} = \sum_{j=1}^M \mathbb{D}_\text{KL}\left[P_\theta (c|X_j)||P_{\theta_\text{orig}}(c|X_j)\right]
    \label{eq:retainloss}
\end{equation}
where $P_{\theta_\text{orig}}$ is the distribution over the choices generated by the original LLM. The overall loss is then defined in Eqn. \eqref{eq:loss}.
\begin{equation}
    \mathcal{L} = \mathcal{L}_\text{unlearn} + \mathcal{L}_\text{retain}
    \label{eq:loss}
\end{equation}
In each minibatch, equal number of unlearning MCQs and retain set MCQs are sampled.

\section{Probing Question Generation}
\label{sec:probing}

\begin{figure*}[t]
    \centering
    \includegraphics[width=0.95\linewidth]{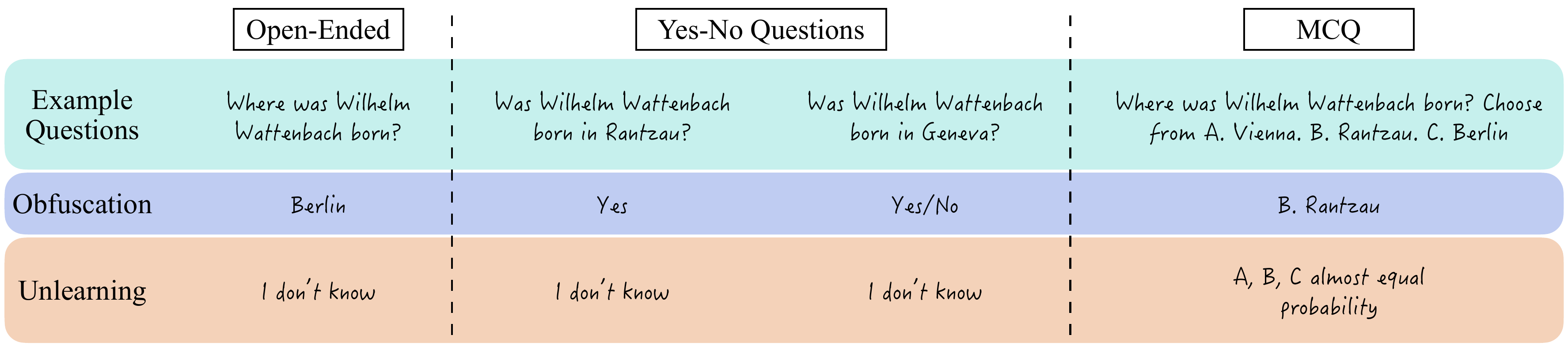}
    \caption{Probing questions to distinguish between obfuscation and unlearning. Open-ended questions (Left) are commonly used in unlearning benchmarks. Yes-No questions (Middle) directly test existence of a connection. Since obfuscation does not remove connections, model is expected to respond yes to the correct answer. Model may respond yes to other questions depending on whether a new connection is established. For MCQ (Right), obfuscation model still has high probability to find correct choice since the connection still exists.}
    \label{fig:probing}
\end{figure*}

This section introduces how we design probing questions to examine whether the effect of a method is unlearning or obfuscation. We group probing questions into three types: (i) \textit{open-ended questions}, (ii) \textit{Yes-No questions} and (iii) \textit{MCQ}. Examples of each type and expected method behaviours are provided in Fig. \ref{fig:probing}.

\subsection{Open-ended Questions}
This is the most commonly used type of questions in unlearning benchmarks such as WPU and TOFU~\cite{tofu}. Obfuscation will cause models to respond with arbitrary answers based on connections built during training. In contrast, unlearning effect should have clear indication that the model does not have information, since there is no existing connections found in model knowledge. Existing evaluation metrics, such as ROUGE-L and GPT privacy scores \cite{whpplus}, will give good performance indications for both obfuscation and unlearning since they both provide answers different to the reference. However, these evaluation metrics are unable to determine whether the knowledge is removed or being obfuscated, and hence motivate us to design the other two types of questions to probe and understand what actually happens to the model knowledge.

\subsection{Yes-No Questions}
This type of questions directly probes whether a connection (hence relevant knowledge) exists or not. For each possible answer, the original open-ended questions is reformulated as one asking whether the answer is correct or not, as shown by two examples in Fig. \ref{fig:probing}. When the connection exists, the model will respond with a certain answer (either yes or no), and when the model truly unlearns it so that the connection does not exist, the model should give a highly uncertain prediction or just respond with ``I do not know". 

To analyze the effect of obfuscation in detail, we split the possible answers to the original open-ended questions into 3 sets based on their sources:

\vspace{-0.2cm}
\begin{itemize}[leftmargin=*]
\setlength\itemsep{-0.3em}
    \item \textbf{Reference Set}: The answers are the ground-truth answers to the original question.
    \item \textbf{In-training Set}: The answers are wrong, but are in the training samples to obfuscate the model.
    \item \textbf{Out-of-training Set}: The answers are wrong and are not in the training samples.
\end{itemize}

\noindent In addition to accuracy in each set, we measure the entropy of predicting \textit{yes} or \textit{no} as follows.
\begin{equation}
    H(X) = -\sum_{y\in\{\text{yes}, \text{no}\}} \bar{P}(y|X) \log \bar{P}(y|X)
\end{equation}
where $X$ is the Yes-No question and $\bar{P}(y|X)$ is the normalized LLM output probability such that $\bar{P}(\text{yes}|X)+\bar{P}(\text{no}|X)=1$.

\subsection{MCQ}
The last type of questions is MCQ as shown on the right side of Fig. \ref{fig:probing}. Instead of asking the model to directly answer the open-ended question, we provide $C$ choices to the model and ask it to choose one from them. Specifically, we use the answer from the \textit{reference set} and one answer from the \textit{in-training set} as two choices, and then fill the rest choices with out-of-training answers.

The performance is measured by multiple choice accuracy as well as the entropy over all choice letters as defined below.
\begin{equation}
    H(X) = - \sum_{c\in\mathcal{C}} \bar{P}(c|X)\ln \bar{P}(c|X)
\end{equation}
where $c$ denotes the token of the letter corresponding to each choice and $\bar{P}(c|X)$ is the normalized LLM output probability such that $\sum_{c\in\mathcal{C}}\bar{P}(c|X)=1$. As a result, we expect the obfuscated model to assign much higher probabilities to the reference and in-training choices as they are concrete edges on its internal knowledge graph.

\section{Experimental Setup}
\subsection{Data Specification}

We focus on the task of privacy protection by forgetting information about individuals, and leverage the Wikipedia Person Unlearning (WPU) \cite{whpplus} benchmark forget-2 set as our main evaluation data. There are five subsets in the forget-2 set, where each subset contains two people to forget. The model is trained to unlearn each subset at one time. Any results reported in this paper are averaged across 5 subsets. To test unlearning efficacy, Yes-No and MCQ probing questions are derived from WPU, in conjunction with the open-ended questions already in the original benchmark. The statistics of different partitions of the probing questions are shown in Table \ref{tab:stats}.

\begin{table}[t]
    \centering
    \small
    \begin{tabular}{lc}
    \toprule
    Dataset Split ~~~~~~~~~~~~~~~ & Number of Questions \\
    \midrule
    \rowcolor{Gray} \multicolumn{2}{l}{Yes-No Questions}  \\
    Reference set & 23 \\
    In-training set & 291 \\
    Out-of-training set & 231 \\
    Retain set & 100 \\
    Hard retain set & 183 \\
    \rowcolor{Gray} \multicolumn{2}{l}{Multiple Choice Questions}  \\
    Forget set & 238 \\
    Hard retain set & 364  \\
    \bottomrule
    \end{tabular}
    \caption{Number of questions on each split of the Yes-No and MCQ probing question sets.}
    \label{tab:stats}
\end{table}

Meanwhile, a retain set containing 100 people is used to measure the performance on people that are not intended to unlearn. Note that this set contains different people from the retrain set used during training. In addition, each subset is associated with a hard retain set containing questions about the target Wikipedia passage that are irrelevant to the target personal information. A good unlearning method should retain the same performance on the retain sets. Probing questions are also created for the hard retain sets. Model performance for Yes-No and MCQ probing questions is evaluated using accuracy and entropy of the model output distribution, and for open-ended questions, ROUGE-L recall is used with the true answer as the reference, following \citet{whpplus}.

To create MCQ training data for DF-MCQ, 20 passages about the person to forget are sampled from the LLM to ensure coverage, and MCQs are generated by prompting the LLM with each generated passage. This yields 300-400 questions for each person. Note that we do not need the correct answer for that MCQ since the goal is to flatten whatever distribution the model predicts. In addition, retain set MCQs are also generated for training following the same procedure for 100 celebrities that do not overlap with WPU forget-2 set. The generation process takes around 10 minutes per target person on a single A100 GPU. 

\subsection{Model and Training}

We use Llama-3.1-8B-Instruct as the main model for evaluation, and demonstrate the generalizability of the properties of DF-MCQ on Qwen-2.5-7B-Instruct. Both models are fine-tuned with low-rank adaptation (LoRA). We choose \textbf{NPO} \cite{npo} and \textbf{WHP$^+$} \cite{whpplus} as two obfuscation methods for comparison with DF-MCQ following their respective implementations. Specifically, WHP$^+$ achieves obfuscation via a model distillation mechanism where the teacher model generates passages about irrelevant individuals, together with per-token distributions of each passage. Then, the names in those passages are replaced by the name of the target person to form obfuscation samples to train the student model.

For DF-MCQ, the model is trained for 3 epochs, which takes 15 minutes on an A100 GPU for each 2-person set. We prompt the LLM to generate a passage about the target person at the end of each epoch, and the learning rate is adjusted such that the model refuses to answer and respond with ``I do not have information".

\section{Results}
\subsection{Open-Ended Questions}

To begin with, the performance of different methods is compared on the standard WPU open-ended questions, and the results are shown in Table \ref{tab:fullopen}. In addition to the ROUGE-L scores, we measure refusal rate as the percentage of questions where LLM responds with no information since this is the expected behaviour of unlearning.

\begin{table*}[t]
    \centering
    \begin{tabular}{lccccc}
    \toprule
    Methods     & Forget Set ($\downarrow$) & Retain Set ($\uparrow$) & Hard Retain Set ($\uparrow$) & Refusal Rate ($\uparrow$) \\
    \midrule
    Original Model & 53.04 & \textbf{91.17} & 59.62 & 0.00 \\
    \midrule
    NPO & 35.23 & 76.85 & 53.85 & 0.00 \\
    WHP$^+$ & 21.01 & 90.12 & 55.65 & 9.23 \\
    DF-MCQ & \textbf{10.70} & 90.34 & \textbf{60.53} & \textbf{92.72} \\
    \bottomrule
    \end{tabular}
    \caption{Performance comparison of NPO, WHP$^+$ and DF-MCQ on open-ended questions from WPU using Llama-3.1-8B-Instruct. Forget set, retain set and hard retain set performance are measured by ROUGE-L recall. The refusal rate is the percentage of responses that refuses to answer questions in the forget set.}
    \label{tab:fullopen}
\end{table*}

\textbf{Main Results}: Overall, DF-MCQ outperforms NPO and WHP$^+$ across all three sets and achieving a refusal rate of \textbf{92.72\%}. NPO on this task significantly degrades the model usability, resulting in a low performance on the two retain sets and hence is excluded for future comparisons. WHP$^+$ is a much more effective obfuscation method compared to NPO for privacy protection without degrading the model performance on the retain set. However, as the model answers the question with an incorrect answer, there are inevitable overlap against the reference answer (e.g. both repeating part of the question), hence not yielding a lower ROUGE-L on the forget set. In contrast, DF-MCQ almost always refuses to answer the question, hence minimizing the possibility of text overlapping and yielding the lowest ROUGE-L among counterparts.

\begin{figure}[h]
    \centering
    \includegraphics[width=1.0\linewidth]{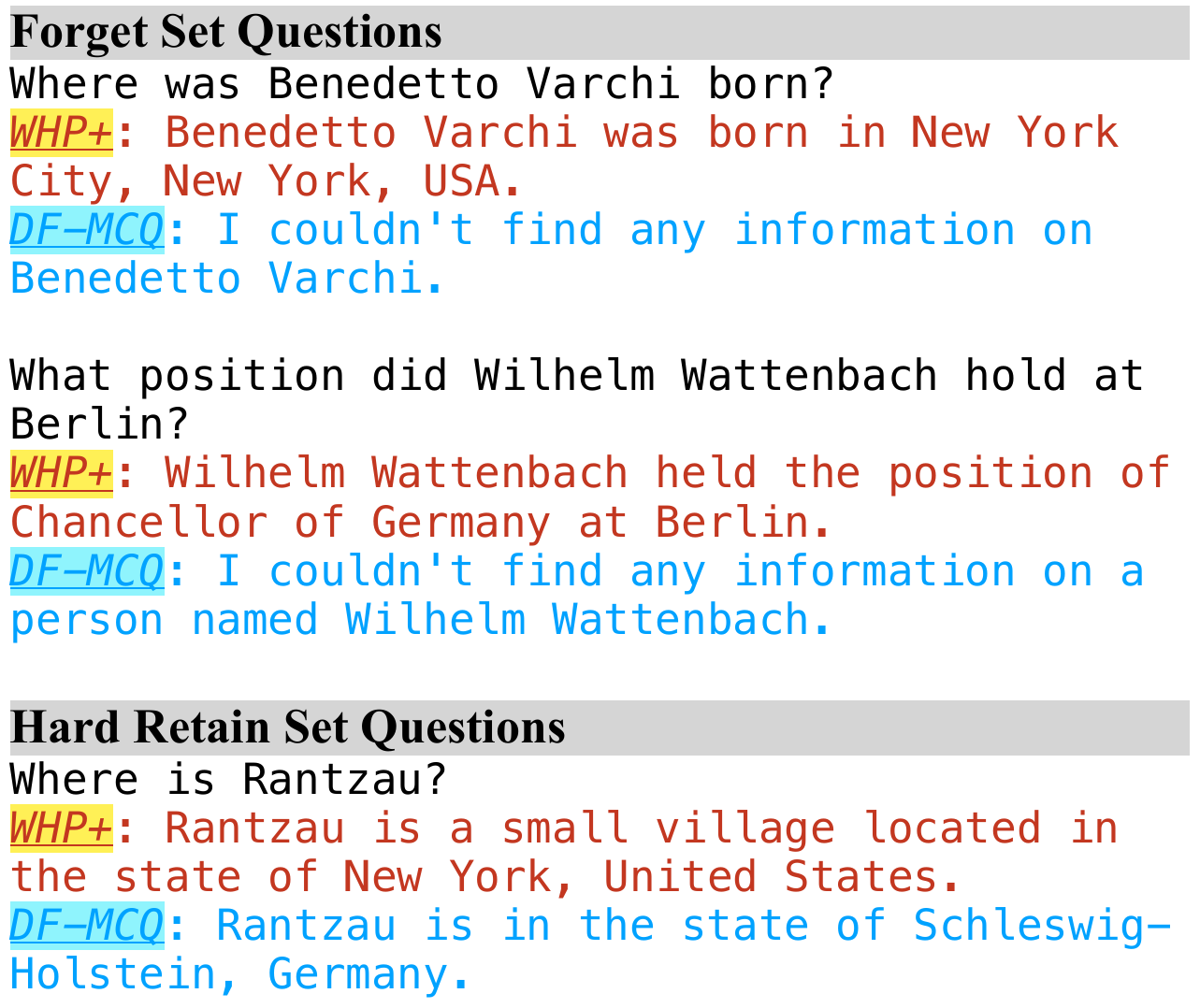}
    \caption{Examples from the forget set and the hard retain set showing responses from LLM trained with WHP$^+$ and DF-MCQ.}
    \label{fig:casestudy}
\end{figure}

\textbf{Case Study}: We use examples in Fig. \ref{fig:casestudy} to further {illustrate the effect of unlearning as apposed to obfuscation}. WHP$^+$ often tries to provide an incorrect answer, likely to be one derived from the teacher model-generated passages. On the contrary, the model trained with DF-MCQ refuses to answer by stating no information found. Another potential problem with obfuscation is the possibility of introducing false edges on the knowledge graph, such as the example shown for the hard retain set. In this example, United State is used in one obfuscation sample to replace the functionality of Germany, causing the model to build an additional wrong connection. This explains why DF-MCQ achieves a slightly better performance on the retain sets. 

\begin{table}[h]
    \centering
    \footnotesize
    \begin{tabular}{lccc}
    \toprule
    Methods     & Forget ($\downarrow$) & Hard Retain ($\uparrow$) & Refusal ($\uparrow$) \\
    \midrule
    Orig. & 51.86 & 60.71 & 0.00 \\
    \midrule
    WHP$^+$ & 28.23 & 57.67 & 12.26 \\
    DF-MCQ & \textbf{17.48} & \textbf{59.53} & \textbf{88.17} \\
    \bottomrule
    \end{tabular}
    \caption{WHP$^+$ and DF-MCQ on open-ended questions from WPU using Qwen2.5-7B-Instruct. Forget set and hard retain set performance are measured by ROUGE-L recall. The refusal rate is the percentage of responses that refuses to answer questions in the forget set.}
    \label{tab:fullopenqwen}
\end{table}

\textbf{Qwen2.5-7B-Instruct Results}: To further validate our observations about refusal, we conduct another set of {experiments on the Qwen2.5-7B-Instruct model}, and the results are shown in Table \ref{tab:fullopenqwen}. We observed similar performance on the forget and hard retain sets, as well as the refusal behaviour, showcasing the generalization of DF-MCQ as an unlearning method across different foundation models. However, Qwen2.5 requires a higher LoRA rank (i.e., more trainable parameters) in order to achieve the desired refusal behaviour.

\textbf{Discussion}: We believe there is no clear boundary between obfuscation and unlearning. This is reflected by the non-zero refusal rate of WHP$^+$. When infinite obfuscation samples are used and the model is updated by seeing enough samples, it achieves knowledge removal. In this case, removing the existing edge is a much easier way than memorizing all possible edges to achieve the flat distribution over the entire output space. However, this is infeasible to achieve as the output space is extremely large for open-ended questions. The DF-MCQ, on the other hand, restricts the output space to only the finite set of choices, where the sum of the probabilities of all choice letters is very close to 1. Therefore, flattening the distribution over the choices is effectively flattening the entire output space, and hence the easiest learning path is to remove the knowledge.

\textbf{Robustness to SFT}: We show the robustness of models unlearnt with DF-MCQ to the supervised fine-tuning (SFT) attack with questions about other individuals in Table \ref{tab:sftattack}. As a result, SFT attack does not influence the forget set performance, and the ones that output "I don't know" still outputs "I don't know" after finetuning.
\begin{table}[h]
    \centering
    \footnotesize
    \begin{tabular}{lcc}
    \toprule
    Model & Forget & Hard Retain \\
    \midrule
    DF-MCQ & 10.70 &	60.53 \\
     + SFT Attack \cite{hu2025unlearningobfuscatingjoggingmemory}	& 10.70 &	61.33 \\
    \bottomrule
    \end{tabular}
    \caption{Robustness to SFT attack on WPU test set.}
    \label{tab:sftattack}
\end{table}

\textbf{Continual Unlearning}: The DF-MCQ can be applied as a continual unlearning method that we can continue adding new unlearning targets without affecting previously unlearnt targets. In theory, this method can work for any number of targets. The results of continually unlearn the 10 individuals, compared to the performance of unlearn just a pair, are shown in Table \ref{tab:continual}.

\begin{table}[h]
    \centering
    \footnotesize
    \begin{tabular}{lccc}
    \toprule
    Model & Forget & Hard Retain & Refusal \\
    \midrule
    DF-MCQ Forget 2	& 10.70	& 60.53	& 92.72 \\
    DF-MCQ Forget 10 & 11.70 & 58.77 & 91.81 \\
    \bottomrule
    \end{tabular}
    \caption{Continually unlearn 10 (Forget 10) individuals comapred to the average performance of the standard unlearn 2 setting in Table \ref{tab:fullopen} (Forget 2).}
    \label{tab:continual}
\end{table}

\subsection{Yes-No Probing Questions}

Then, Yes-No probing questions are used to further analyze obfuscation and unlearning effects, where the results are shown in Table \ref{tab:yesno}. Since DF-MCQ tends to refuse to answer, we add ``You must answer Yes or No" to the prompt to force it respond.

\begin{table*}[t]
    \centering
    \begin{tabular}{lccccc}
    \toprule
    Methods     & Reference & In-training & Out-of-training & Retain & Hard Retain \\
    \midrule
    Original Model     &  100.0 (0.09) &	69.56 (0.29) & 52.28 (0.26)	& 56.57 (0.28) & 45.00 (0.41) \\
    \midrule
    WHP$^+$ & 100.0 (0.43) & 0.0 (0.46) & 0.0 (0.48) & 29.95 (0.47) & 24.30 (0.49) \\
    DF-MCQ     & 77.60 (0.65) & 41.90 (0.66) & 34.46 (0.64) & 52.87 (0.31) & 37.34 (0.44) \\
    \bottomrule
    \end{tabular}
    \caption{Accuracies and entropy (in bracket) on the three separate test sets of Yes-No questions as well as the retain set and the hard retain set. The maximum entropy for binary output is 0.69 with natural log. The correct answer for the reference set is always ``Yes", and that for the in-training and out-of-training sets is always ``No".}
    \label{tab:yesno}
\end{table*}


\textbf{Main Results}: The expected behaviour of unlearning is that the model does not have knowledge about the person, which corresponds to high entropy when answering these Yes-No probing questions. While WHP$^+$ increases the entropy of model prediction on the reference set, it fails to reduce the accuracy, whereas DF-MCQ largely reduces the accuracy and achieves an entropy of \textbf{0.65}, close to a random guess. Moreover, for WHP$^+$, obfuscation causes the model to find the shortcut that always answers Yes whenever it sees the target name appear in the prompt. Since the reference answers of the in-training and out-of-training sets are always ``No'', WHP$^+$ yields zero accuracy on those sets. As before, DF-MCQ achieves high entropy, indicating that the model truly does not know the answer.

The retain sets do not contain the target names and hence the obfuscation model does not always respond ``Yes" to the questions. This suggests that the shortcut behaviour is mainly tied to the target names rather than the question type. Nevertheless, the performance of WHP$^+$ still degrades on those sets due to unintended edges established during training. In contrast, DF-MCQ achieves much better accuracy than the obfuscation method, and in particular, achieves the same level of uncertainty to the original model on the two retain sets.

\textbf{Different split for DF-MCQ}: To illustrate that DF-MCQ is not obfuscation by the distracting options, a new split for Yes-No probing questions is adopted. Instead of using the in-training and out-of-training sets derived from the obfuscation passages, we treat the distracting choices in the training set MCQs as the in-training set.

As a result, DF-MCQ achieved 25.47\% accuracy on the new in-training set with entropy of 0.63, and an accuracy of 30.56\% with entropy of 0.64 on the new out-of-training set. This indicates that for any questions regarding the target person, no matter whether it corresponds to a choice in the training set or not, the model behaviour is always close to a random guess, with some inevitable priors, e.g. names may suggest nationalities. Therefore, DF-MCQ removes knowledge and is clearly different from obfuscation methods.

\begin{figure}[t]
    \centering
    \includegraphics[width=\linewidth]{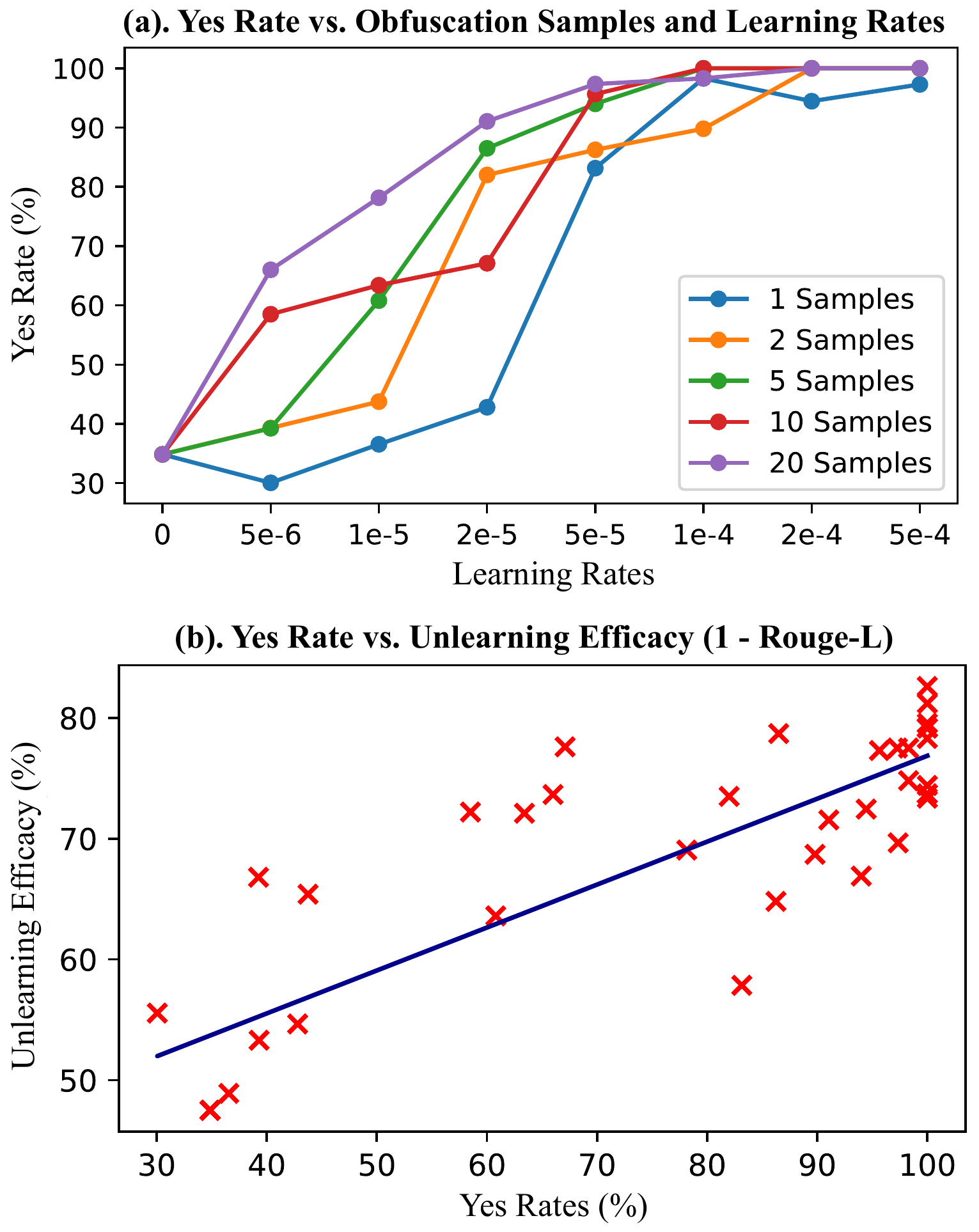}
    \caption{The rate of answering Yes (Yes rate) against learning rates and number of obfuscation samples (a) and correlation between unlearning efficacy and Yes rate (b). Each point in (b) corresponds to a point in (a) with unlearning efficacy measured by $1-$ROUGE-L. The Pearson Correlation Coefficient of (b) is 0.84.}
    \label{fig:yesnoplot}
    \vspace{-0.3cm}
\end{figure}

\textbf{Shortcut to always answer Yes}: We investigate this shortcut behaviour of the obfuscation method against the number of obfuscation samples and learning rate, and plot the rate of answering Yes (Yes rate) as shown in Fig. \ref{fig:yesnoplot}(a). Since the in-training set changes with the obfuscation samples used, and according to Table \ref{tab:yesno} this shortcut behaviour is agnostic to the split, we measure the Yes rate on the same out-of-training set.

First, increasing the number of obfuscation samples increases the tendency of shortcut. Second, with an increasing learning rate and hence larger model updates, the model is more likely to answer Yes. We also plot the correlation between unlearning efficacy measured by $1-$ROUGE-L on the open-ended questions and the Yes rate as shown in Fig. \ref{fig:yesnoplot} (b). The Pearson Correlation Coefficient is 0.84. That is, \textit{to obfuscate the model to a degree that effectively protects privacy, the model is very likely to answer Yes to all probing questions about the target person}.

\subsection{MCQ Probing Questions}

\begin{table}[t]
    \centering
    \footnotesize
    \begin{tabular}{lccccc}
    \toprule
    Methods     & \multicolumn{3}{c}{Forget} & \multicolumn{2}{c}{Hard Retain} \\
    & Acc. & $H$ & $P(c_\text{obf.})$ & Acc. & $H$ \\
    \midrule
    Orig. Model  & 74.26 & 0.18 & 0.03 & 76.73 & 0.20 \\
    \midrule
    WHP$^+$  & 36.73 & 1.10 & 0.29 & 73.08 & 0.55 \\
    DF-MCQ     & 18.86 & 1.61 & 0.20 & 63.19 & 0.66 \\
    \bottomrule
    \end{tabular}
    \caption{Accuracies, entropy and probability of obfuscation choices ($P(c_\text{obf.})$) on the forget and hard retain sets of MCQs using different methods. The maximum entropy for 5 choices is 1.61 with natural log.}
    \label{tab:mcqprobe}
    \vspace{-0.3cm}
\end{table}

The last part of the experiments uses MCQ as probing questions to evaluate the behaviour of obfuscation versus unlearning. Results are reported on the forget set and the hard retain set, as shown in Table \ref{tab:mcqprobe}. In addition to the accuracy and entropy as before, for the forget set questions, we measure the probability of the obfuscation choice, $P(c_\text{obf.})$.

\textbf{Obfuscation has limited efficacy with MCQ}: While obfuscation can drive the model to give wrong answers for open-ended questions, as it does not remove the knowledge and when edges of other choices are not established, it still has a tendency to choose the correct answer for MCQs. As a result, WHP$^+$ has a reasonably high accuracy of 36.73 on the forget set, and in particular, close to the original model performance for a couple of target individuals (see Appendix \ref{sec:breakdown} for breakdown results on subsets). This indicates that when a set of candidates are presented to the obfuscation model, it may fail to protect privacy.

The obfuscation effect also raises the likelihood of selecting the option that appeared in the obfuscation samples used during training, as indicated by $P(c_\text{obf.})$ in Table \ref{tab:mcqprobe}. On the contrary, DF-MCQ assigns almost equal probability to all options, subject to certain priors, and it is impossible to infer which information was used during unlearning. Therefore, compared to obfuscation, DF-MCQ better protects the privacy when a malicious query contains a range of options.

\textbf{Retain set performance}: Although DF-MCQ trains the model to flatten the output distribution over its choices, this flatten behaviour is mainly tied to the target individual rather than the MCQ question type. This is reflected by the performance on the retain set shown in Table \ref{tab:mcqprobe}. Admittedly, DF-MCQ does have a slight shortcut impact to the accuracy due to the model being exposed to only MCQ tasks, this impact is much smaller compared to the catastrophic shortcut observed in obfuscation method on Yes-No questions.

\section{Conclusions}
We investigate the effect of unlearning from an uncertainty perspective, and propose the distinction between true unlearning and obfuscation. We identify the refusal behaviour of true unlearning effect as apposed to obfuscation effect which provides wrong answers, and propose a set of probing questions to help distinguish the them. Furthermore, DF-MCQ is proposed which achieves true unlearning by flattening the distribution of answers to MCQs. As a result, DF-MCQ achieves over 90\% refusal rate to open-ended questions about the unlearning target, as well as achieving a random choice-level uncertainty that is much higher than obfuscation methods on probing questions.

\section*{Limitations}
This study examines person‑centric facts following WPU and mid‑sized instruction models (under 10B parameters). Extending DF‑MCQ to broader contents, multilingual, or multimodal data could be an interesting future work. Because DF‑MCQ relies on automatically generated multiple‑choice questions, improving distractor diversity and pipelines would further strengthen the method. 

\section*{Acknowledgments}
Guangzhi Sun is supported by junior research fellowship from Trinity College, Cambridge.

\bibliography{anthology, custom}
\bibliographystyle{acl_natbib}

\appendix
\newpage
\section{Break Down Results for MCQ Probing}
\label{sec:breakdown}

We provide breakdown results for MCQ probing questions to show the possible failure mode of obfuscation on specific individuals. The performance of the original model, WHP$^+$ and DF-MCQ are shown in Tables \ref{tab:breakdown1}, \ref{tab:breakdown2} and \ref{tab:breakdown3} respectively.

\begin{table}[h]
    \centering
    \footnotesize
    \begin{tabular}{lccccc}
    \toprule
    Subsets &	Accuracy & Entropy & Prob & Accuracy & Entropy \\
    \midrule
    Set 1& 92.75 & 0.03 & 0.01 & 80.00 & 0.23 \\
    Set 2 &72.50 & 0.28 & 0.11 & 75.64 & 0.10 \\
    Set 3& 57.35 & 0.37 & 0.00 & 70.97 & 0.26 \\
    Set 4& 100.0 & 0.03 & 0.01 & 79.41 & 0.18 \\
    Set 5 & 48.72 & 0.18 & 0.00 & 77.63 & 0.24 \\
    \midrule
    Overall & 74.26	& 0.18	& 0.03 &	76.73 & 0.20 \\
    \bottomrule
    \end{tabular}
    \caption{Breakdown results for 2-person subsets of the original model performance on probing MCQs.}
    \label{tab:breakdown1}
\end{table}

\begin{table}[h]
    \centering
    \footnotesize
    \begin{tabular}{lccccc}
    \toprule
    Subsets &	Accuracy & Entropy & Prob & Accuracy & Entropy \\
    \midrule
    Set 1	& 40.58 &	1.17 &	0.23 &	66.25 &	0.72 \\
    Set 2	& 32.50 &	1.25 & 0.27 & 76.92	& 0.44 \\
    Set 3	& 20.59& 0.96	& 0.46&	72.58	& 0.56 \\
    Set 4	& 18.18& 1.22	 & 0.39&	79.41	& 0.47 \\
    Set 5	& 71.79& 0.92 &	0.11&	70.26	& 0.57 \\
    \midrule
    Overall & 36.73 &	1.10	& 0.29 &	73.08 &	0.55 \\
    \bottomrule
    \end{tabular}
    \caption{Breakdown results for 2-person subsets of WHP$^+$ performance on probing MCQs.}
    \label{tab:breakdown2}
\end{table}

\begin{table}[h]
    \centering
    \footnotesize
    \begin{tabular}{lccccc}
    \toprule
    Subsets &	Accuracy & Entropy & Prob & Accuracy & Entropy \\
    \midrule
Set 1 &	15.94&	1.61	& 0.20 &	67.50 &	0.71 \\
Set 2&	12.50&	1.61	& 0.20	& 53.85 &	0.42\\
Set 3&	22.06&	1.61	& 0.20	& 58.06	& 0.80\\
Set 4&	18.18&	1.61	& 0.20	& 72.06	& 0.54\\
Set 5&	25.64&	1.61	& 0.20 &	64.47 &	0.85\\
\midrule
	Overall & 18.86 &	1.61 &  0.20	& 63.19 &	0.66 \\
    \bottomrule
    \end{tabular}
    \caption{Breakdown results for 2-person subsets of DF-MCQ performance on probing MCQs.}
    \label{tab:breakdown3}
\end{table}

\end{document}